\documentclass{article}

\usepackage{microtype}
\usepackage{graphicx}
\usepackage{subfigure}
\usepackage{booktabs} % for professional tables
\usepackage{algorithm}
\usepackage{algorithmic}
\usepackage{amsmath,amssymb}
\usepackage{nicefrac}
\usepackage{bm}

\usepackage{hyperref}

\usepackage[accepted]{icml2020}

\icmltitlerunning{SWAG: A Wrapper Method for Sparse Learning}

\begin{document}

\twocolumn[
\icmltitle{SWAG: A Wrapper Method for Sparse Learning}



\icmlsetsymbol{equal}{*}

\begin{icmlauthorlist}
\centering
\icmlauthor{Roberto Molinari\textsuperscript{*}}{au}
\icmlauthor{Gaetan Bakalli\textsuperscript{*}}{au}
\icmlauthor{St\'{e}phane Guerrier}{ug}
\icmlauthor{Cesare Miglioli}{ug}
\icmlauthor{Samuel Orso}{ug}
\icmlauthor{Mucyo Karemera}{au}
\icmlauthor{Olivier Scaillet}{ugs}
\end{icmlauthorlist}

\icmlaffiliation{au}{Department of Mathematics and Statistics, Auburn University, Auburn, AL 36849, USA}
\icmlaffiliation{ug}{Geneva School of Economics and Management, University of Geneva, Geneva, 1205, Switzerland}
\icmlaffiliation{ugs}{Geneva School of Economics and Management, University of Geneva, Geneva, 1205, Switzerland and Swiss Finance Institute}

%\icmlcorrespondingauthor{}{}

\icmlkeywords{Machine Learning, ICML}

\vskip 0.3in
]



\printAffiliationsAndNotice{\icmlEqualContribution} % otherwise use the standard text.

\begin{abstract}
The majority of machine learning methods and algorithms give high priority to prediction performance which may not always correspond to the priority of the users. In many cases, practitioners and researchers in different fields, going from engineering to genetics, require interpretability and replicability of the results especially in settings where, for example, not all attributes may be available to them. As a consequence, there is the need to make the outputs of machine learning algorithms more interpretable and to deliver a library of “equivalent” learners (in terms of prediction performance) that users can select based on attribute availability in order to test and/or make use of these learners for predictive/diagnostic purposes. To address these needs, we propose to study a procedure that combines screening and wrapper approaches which, based on a user-specified learning method, greedily explores the attribute space to find a library of sparse learners with consequent low data collection and storage costs. This new method (i) delivers a low-dimensional network of attributes that can be easily interpreted and (ii) increases the potential replicability of results based on the diversity of attribute combinations defining strong learners with equivalent predictive power. We call this algorithm ``Sparse Wrapper AlGorithm'' (SWAG).
\end{abstract}

\section{Motivation}
\label{sec.motivation}

The purpose of any machine learning algorithm is to deliver precise predictions with respect to a response (or responses) of interest, whether dealing with classification or more general regression problems. Given this common goal, the focus of research has been in developing algorithms and methods that make use of attribute transformation, filtering and selection techniques that allow to best predict the response through often complex non-linear functions and/or heuristic procedures. This is the case of simple classifiers such as Support Vector Machines (SVM) or methods such as neural networks and deep learning (with all their adaptations). As a result of these developments it is possible to train these algorithms to obtain highly accurate testing and validation predictions with consequent important gains in terms of decision-making effectiveness in all domains in which such techniques are employed. Nevertheless, while prediction accuracy remains a paramount goal, there has been an emerging push in different fields to obtain ``interpretable'' predictions \citep[see e.g.][]{wang2019gaining} in order to better understand phenomena and direct future avenues of research. An important area of machine learning research that (indirectly and partially) addresses this need is that of \textit{sparse} learning \citep[see e.g.][]{chandrashekar2014survey,zhang2015survey} which, in the context of this work, refers to the use of learners that select and make use only of a reduced number of attributes in a dataset (as opposed to all of them). Indeed, aside from guaranteeing reduced computational complexity for prediction when employing fewer attributes and therefore reducing costs for data collection and storage, sparse learners are effective in addressing the common problem of overfitting by excluding attributes that are possibly not informative and that can increase prediction variability. As a consequence, by selecting and making use of a small set of attributes, sparse learners can also lend themselves to being more easily interpreted and can consequently be used a basis to further investigate certain phenomena \citep[for an overview see][]{chandrashekar2014survey}.

However, despite their predictive advantages, all the current sparse learning mechanisms select a single learner (with corresponding unique attributes) and, in high-dimensional settings, nevertheless tend to select many attributes especially in a highly correlated environment \citep[see e.g. ][]{meinshausen2009lasso, vats2013doubt}. These features can have important practical impacts in different settings where interpretability and replicability of the learners is of relevance \citep[see e.g.][]{marigorta2018replicability, quinn2020deepcoda}. Indeed, from genomics \citep[e.g.][]{xiong2001biomarker} to online prediction \citep[e.g.][]{carmona2011dataminingonline}, there are many tasks that require a degree of flexibility in the use of a multitude of (small) subsets of attributes while preserving high predictive performance. For example, (i) in medical studies machines collect different measurements (attributes) for a specific problem \citep[e.g.][]{draghici2006reliability}; (ii) for online search algorithms every subject provides different attributes (according to their preferences or willingness to disclose information) to determine suggestions or matches \citep[e.g.][]{vaughan2015data} ; (iii) in pattern recognition, images are collected at different resolutions and therefore a single learner may not be flexible enough to adapt to different image features \citep[e.g.][]{wang2018interactive}. In addition, interpretability of phenomena can be greatly enhanced when considering a multitude (library) of sparse learners as highlighted by the multimodel inference and model selection uncertainty literature in areas such as sociology and, especially, ecology \citep[see e.g.][for an overview]{burnham2004multimodel, anderson2004model, harrison2018brief}. Moreover, the idea of having a library of learners (and possibly of attributes) was considered, for example, in \cite{caruana2004ensemble} where libraries are created by forward-selection based on ensemble prediction. 

\begin{figure*}[!ht]
\tiny
\begin{minipage}[c]{0.50\linewidth}
\centering
\includegraphics[width=\linewidth]{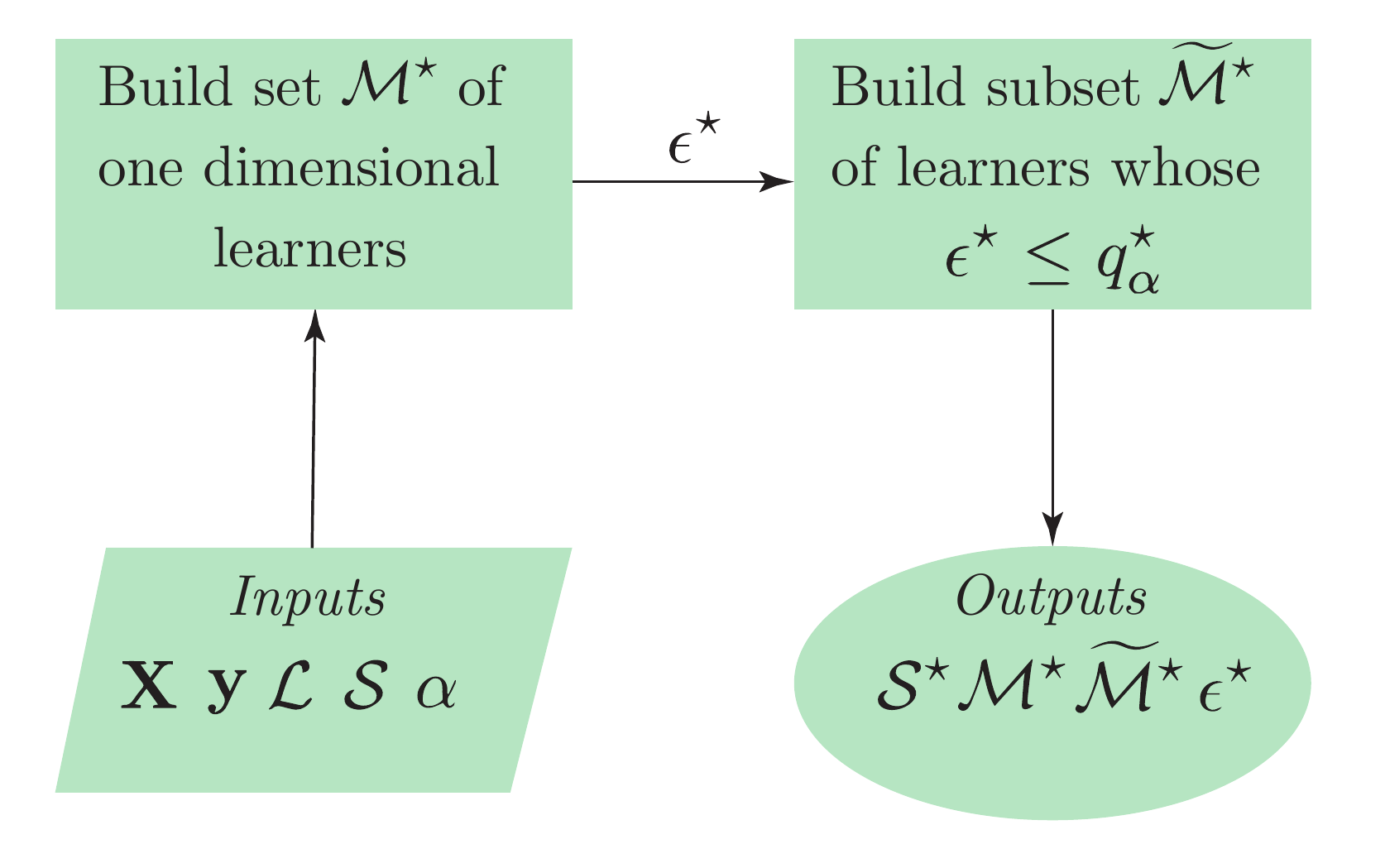}
\end{minipage}
\hspace{.3cm}
\begin{minipage}[c]{0.475\linewidth}
\centering
\begin{algorithm}[H]\small
\caption{First Screening Algorithm}
\scriptsize
  \vspace{0.1cm}
INPUTS: A response $\mathbf{y} \in \mathbb{R}^n$ and attributes $\mathbf{X} \in \mathbb{R}^{n \times p}$; An attribute index set $\mathcal{S} := \{1, \hdots, p\}$; A learning mechanism $\mathcal{L}$; A performance percentile $\alpha \in (0,1)$; the number of repetitions $r$ and the number of folds $k$ to compute the cross-validation training error. \\
\begin{algorithmic}[1]
  \setlength\itemsep{0em}
\STATE Using the mechanism $\mathcal{L}$, build $p$ learners  by using all attributes in the set $\mathcal{S}$
\STATE Create a learner set $\mathcal{M}^{\star}$ (with $|\mathcal{M}^{\star}| = p$)
\STATE Build a an $r$-repeated $k$-fold cross-validation error vector $\bm{\epsilon}^{\star} \in \mathbb{R}^{p}$ and identify the $\alpha$-quantile $q_{\alpha}^{\star}$ of this vector
\STATE Create new learner set $\widetilde{\mathcal{M}}^{\star}$ with learners whose cross-validation error is smaller or equal to $q_{\alpha}^{\star}$
\STATE Create attribute index set $\mathcal{S}^\star$ with attributes included in the learners in the set $\widetilde{\mathcal{M}}^{\star}$.

\end{algorithmic}
  \vspace{0.3cm}
OUTPUTS: $\mathcal{S}^\star$; $\widetilde{\mathcal{M}}^{\star}$; $\mathcal{M}^{\star}$; $\bm{\epsilon}^{\star}$.
\label{algo.firstscreen}
\end{algorithm}
\end{minipage}\hfill
\end{figure*}

Considering the above discussion, this work aims at creating a library of ``strong'' and low-dimensional learners which can be selected either (i) \textit{individually} based on attribute availability or based on whether some attributes are of specific interest within a given domain and/or (ii) \textit{jointly} to generate a low-dimensional network highlighting the intensity and direction of attribute interaction in order to explain the responses of interest. The method we put forward for this purpose consists in a new greedy version of the algorithm that was proposed for gene selection problems in \cite{guerrier2016predictive} and overcomes some of its limitations in terms of dimension reduction. We call this development the ``Sparse Wrapper AlGorithm'' (SWAG for short) whose main goal is to explore the attribute space in an ``informed'' manner to deliver a library of low-dimensional learners, all with high and comparable predictive power. Hence, given a selected learning method, the output of this approach therefore increases the possibility of replicating results and of making use of specific learners of interest while delivering interpretable connections between attributes for possible future directions of research in different domains.

%\section{\textcolor{blue}{Latent variables analysis}}
%\textcolor{blue}{By selecting not one predictive model, but a class of equivalent models, the SWAG opens a gateway on the latent variables that structure the data. The use of latent variables can serve to reduce the dimensionality of data, as observable variables can be aggregated in a model to represent an underlying concept. This process makes it easier to understand and interpret results. In this sense, they serve a function similar to that of scientific theories. In biology, integrating data from multiple global assays and curated databases is essential to understand the spatio-temporal interactions within cells. Different experiments measure cellular processes at various widths and depths, while
%databases contain biological information based on established facts or published data \cite{zhang2008network}. The recent advancement of Omic technologies provides researchers with the possibility to search for disease-associated biomarkers at the system level . The integrative analysis of data from a
%large number of molecules involved at various layers of the biological system offers a great opportunity
%to rank disease biomarker candidates \cite{fan2020mota} .Integrating these complementary
%data sets helps infer consistent relationships among genomic features and build intelligible data-based networks. }

\section{The Sparse Wrapper Algorithm}

The SWAG combines screening approaches within a general wrapper method to explore the (low-dimensional) attribute space in an effective manner. In order to find a library (set) of learners that take a small number of attributes as inputs and that have high predictive power, the algorithm proceeds in a forward-step manner. More specifically, it builds and tests learners starting from very few attributes until it includes a maximal number of attributes by increasing the number of attributes at each step. Hence, for each fixed number of attributes, the algorithm tests various (randomly selected) learners and picks those with the best performance in terms of training error. Throughout, the algorithm uses the information coming from the best learners at the previous step to build and test learners in the following step. In the end, it outputs a set of strong low-dimensional learners.

Given the above intuitive description, we now provide a more formal description and, for this reason, we define some basic notation. Let $\mathbf{y} \in \mathbb{R}^n$ denote the response and $\mathbf{X} \in \mathbb{R}^{n \times p}$ denote an attribute matrix with $n$ instances and $p$ attributes, the latter being indexed by a set $\mathcal{S} := \{1, \hdots, p\}$. In addition, a generic learning mechanism is denoted as $\mathcal{L}:= \mathcal{L}(\mathbf{y}, \mathbf{X})$ with $l$ denoting a general learner which is built by using (i) the learning mechanism $\mathcal{L}$ and (ii) a subset of attributes in $\mathbf{X}$. Finally, we let $\mathcal{P}(\mathcal{A})$ and $|\mathcal{A}|$ denote respectively the power set and cardinality of a set $\mathcal{A}$. In the following paragraphs we will proceed to describing the algorithm and introduce meta-parameters whose interpretation and selection will be discussed later in Sec. \ref{sec.meta_param}.

%Given the goal of creating a set of sparse learners with improved (or at least equivalent) predictive power compared to (sparse) learners that take \textit{all} attributes in $\mathcal{S}$ as an input, 

The first choice to make for the algorithm is to determine the maximum dimension of attributes that the user wants to be input in a learner and we denote this parameter as $p_{\text{max}} < p$. Based on this parameter, the SWAG aims at exploring the space of attributes in order to find sets of learners using $\hat{p}$ attributes ($1 \leq \hat{p} \leq p_{\text{max}}$) with high predictive power. To do so, the algorithm makes use of the step-wise screening procedure described in the following paragraphs.

\paragraph{First Step}
The first screening step starts by using one \textit{distinct} attribute at a time to create $p$ learners. Once these learners are built, a set of learners $\mathcal{M}^{\star}$ is now available which is indexed by the ordered index set $\mathcal{I} := \{1, \hdots, p\}$ (i.e. each learner $l \in \mathcal{M}^{\star}$ is indexed by a unique element $i \in \mathcal{I}$). Having chosen a measure of predictive error (e.g. a loss function such as the misclassification rate for classification problems), one can then apply $r$-repeated $k$-fold cross-validation to determine the training error of each learner $l \in \mathcal{M}^{\star}$. We denote the vector containing the cross-validation error as $\bm{\epsilon}^{\star} \in \mathbb{R}^{p}$ which is also indexed by the set $\mathcal{I}$ (i.e. each learner $l \in \mathcal{M}^{\star}$ is associated with an element in the training error vector $\bm{\epsilon}^{\star}$). Given this, it is now possible to select a performance quantile $q_{\alpha}^{\star}$ where $\alpha \in (0, 1)$ is chosen by the user. This quantile is defined such that the following expression holds true:
    $\nicefrac{1}{p} \sum_{i=1}^{p} \mathbb{I}_{\left\{\bm{\epsilon}_i^{\star} \leq q_{\alpha}^{\star}\right\}} = \alpha $,
where $\mathbb{I}_{A}$ is the indicator function that takes the value $1$ if $A$ is true and $0$ otherwise, while $\bm{\epsilon}_i^{\star}$ denotes the i$^{th}$ element of the vector $\bm{\epsilon}^{\star}$. The smaller the value of $\alpha$, the smaller the training errors selected. The procedure then selects all the learners whose training error is smaller or equal to $q_{\alpha}^{\star}$ and includes these in a new learner set $\widetilde{\mathcal{M}}^{\star}$. The set $\widetilde{\mathcal{M}}^{\star}$ therefore collects one-dimensional attribute learners with high predictive power and therefore also contains a subset of attributes $\mathcal{S}^\star \subset \mathcal{S}$ that can be assumed to be highly informative with respect to the response of interest $\mathbf{y}$. This procedure is described in Algo.~\ref{algo.firstscreen}.

\paragraph{General Step}
Fixing a given attribute dimension $\hat{p}$ such that $2 \leq \hat{p} \leq  p_{\text{max}}$, the general screening step builds a maximum number $m$ of \textit{distinct} learners, all included in a learner set $\mathcal{M}^{\hat{p}}$, that each take combinations of $\hat{p}$ \textit{distinct} attributes. In order to build the $m$ learners, the general step takes the attribute index set $\mathcal{S}^\star$ from Algo.~\ref{algo.firstscreen} and a set of learners $\widehat{\mathcal{M}}$ in which each learner is of dimension $\hat{p} - 1$ (i.e. each learner in $\widehat{\mathcal{M}}$ takes $\hat{p} - 1$ attributes as an input). We let $s_l \in  \tilde{S} := \{s \in \mathcal{P}(\mathcal{S}^\star) \,|\, |s| = \hat{p} - 1\}$ denote the attribute indices for a specific learner $l \in \widehat{\mathcal{M}}$. 

In order to build the $m$ learners, the procedure first verifies whether all possible attribute combinations can be used. Indeed, if we denote $p^\star := |\mathcal{S}^\star|$, then the total number of learners that can be built is $\tilde{m} := \binom{p^\star}{\hat{p}}$ and, if $m \geq \tilde{m}$, then the learner set $\mathcal{M}^{\hat{p}}$ contains all possible learners. However, if $m < \tilde{m}$, then this step randomly selects a learner $l \in \widehat{\mathcal{M}}$ and builds a new learner by using the attributes indexed by $s_l$ and a randomly selected distinct attribute from the attribute index set $\mathcal{S}^\star / s_l$. This step is repeated until $m$ distinct learners are built and included in the learner set $\mathcal{M}^{\hat{p}}$.

\begin{figure*}[!ht]
\tiny
\begin{minipage}[c]{0.475\linewidth}
\centering
\begin{algorithm}[H]\small
\caption{General Screening Algorithm}
\scriptsize
  \vspace{0.1cm}
INPUTS: A response $\mathbf{y} \in \mathbb{R}^n$ and attributes $\mathbf{X} \in \mathbb{R}^{n \times p}$; An attribute index set $\mathcal{S}^\star \subset \{1, \hdots, p\}$ from Algo. \ref{algo.firstscreen}; A number of attributes $\hat{p} \leq p_{\text{max}}$; A learner set $\widehat{\mathcal{M}}$; A learning mechanism $\mathcal{L}$; A maximum number of learners $m$; A performance percentile $\alpha \in (0,1)$; the number of repetitions $r$ and the number of folds $k$ to compute the cross-validation training error. \\
\begin{algorithmic}[1]
  \setlength\itemsep{0em}
\STATE Define $\tilde{m} := \binom{p^\star}{\hat{p}}$
\IF {$\tilde{m} \leq m$}
    \STATE Using the mechanism $\mathcal{L}$, build all possible $\tilde{m}$ learners with $\hat{p}$ attribute inputs to create learner set $\mathcal{M}^{\hat{p}}$
\ELSE
    \STATE Using the mechanism $\mathcal{L}$, build $m$ learners with $\hat{p}$ attribute inputs by extracting $s_l$ from randomly sampled learners $l \in \widehat{\mathcal{M}}$ and adding an attribute from $\mathcal{S}^\star / s_l$ to create learner set $\mathcal{M}^{\hat{p}}$
\ENDIF
\STATE Build an $r$-repeated $k$-fold cross-validation error vector $\bm{\epsilon}^{\hat{p}}$ and identify the $\alpha$-quantile $q_{\alpha}^{\hat{p}}$ of this vector
\STATE Create new learner set $\widetilde{\mathcal{M}}^{\hat{p}}$ with learners whose cross-validation error is smaller or equal to $q_{\alpha}^{\hat{p}}$

\end{algorithmic}
  \vspace{0.3cm}
OUTPUTS: $\widetilde{\mathcal{M}}^{\hat{p}}$; $\mathcal{M}^{\hat{p}}$; $\bm{\epsilon}^{\hat{p}}$.
\label{algo.generalscreen}
\end{algorithm}
\end{minipage}
\hspace{.6cm}
\begin{minipage}[c]{.495\linewidth}
\centering
\includegraphics[width=\linewidth]{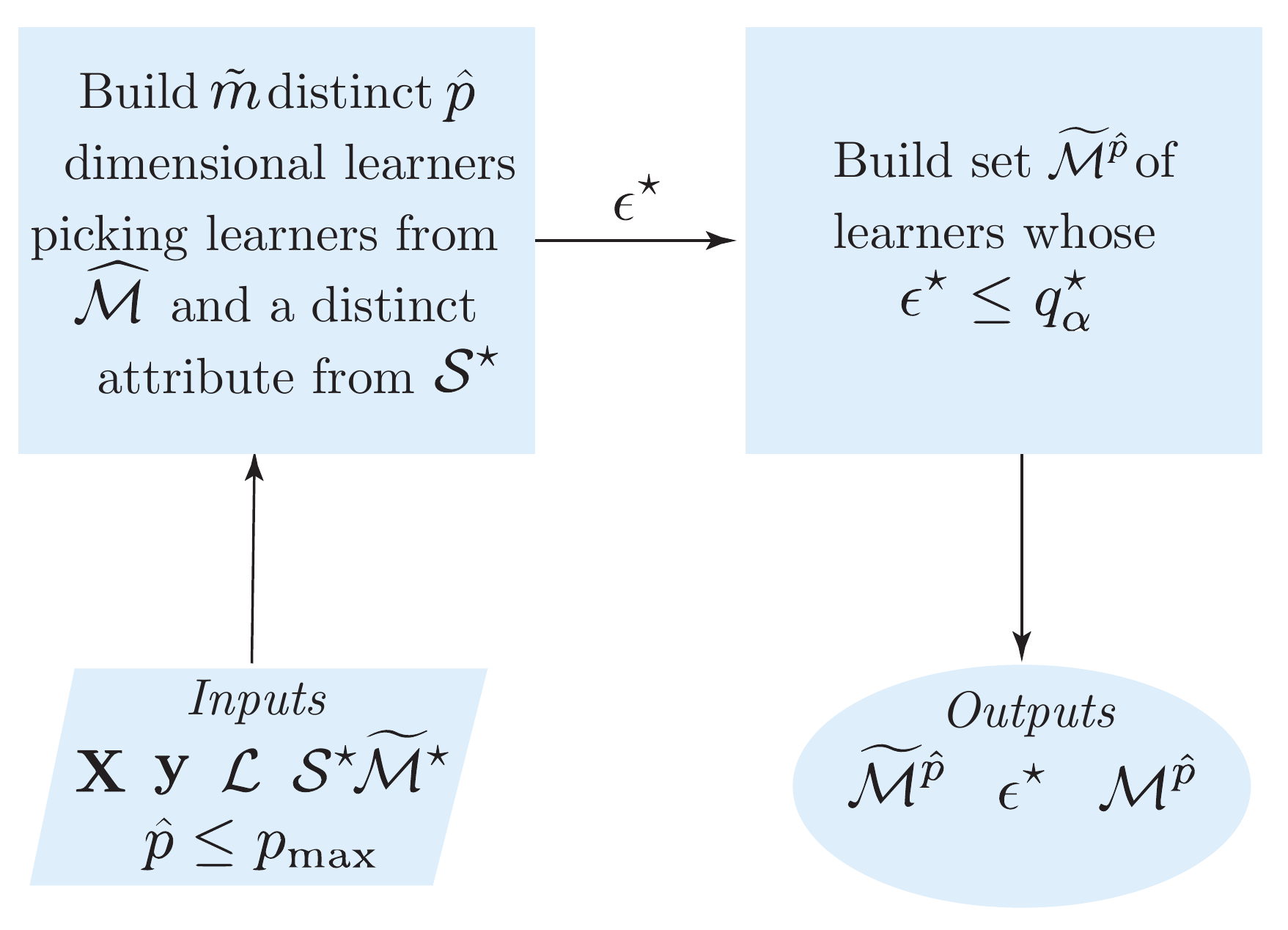}
\end{minipage}\hfill
\end{figure*}

Once the candidate learner set $\mathcal{M}^{\hat{p}}$ is built, the general step of the algorithm closely follows Algo.~\ref{algo.firstscreen}. More specifically, an $r$-repeated $k$-fold cross-validation error $\bm{\epsilon}^{\hat{p}}$ is built and its performance quantile $q_{\alpha}^{\hat{p}}$ is computed. The main output of this step is then a new learner set $\widetilde{\mathcal{M}}^{\hat{p}}$ which includes all learners whose training error in $\bm{\epsilon}^{\hat{p}}$ is smaller or equal to $q_{\alpha}^{\hat{p}}$ (with both $\bm{\epsilon}^{\hat{p}}$ and $\mathcal{M}^{\hat{p}}$ being ordered by the same index set $\mathcal{I}^{\hat{p}} := \{1, \hdots, m\}$). This general step is described in Algo.~\ref{algo.generalscreen}.

\begin{figure*}[!ht]
\tiny
\begin{minipage}[c]{0.50\linewidth}
\centering
\includegraphics[scale = .43]{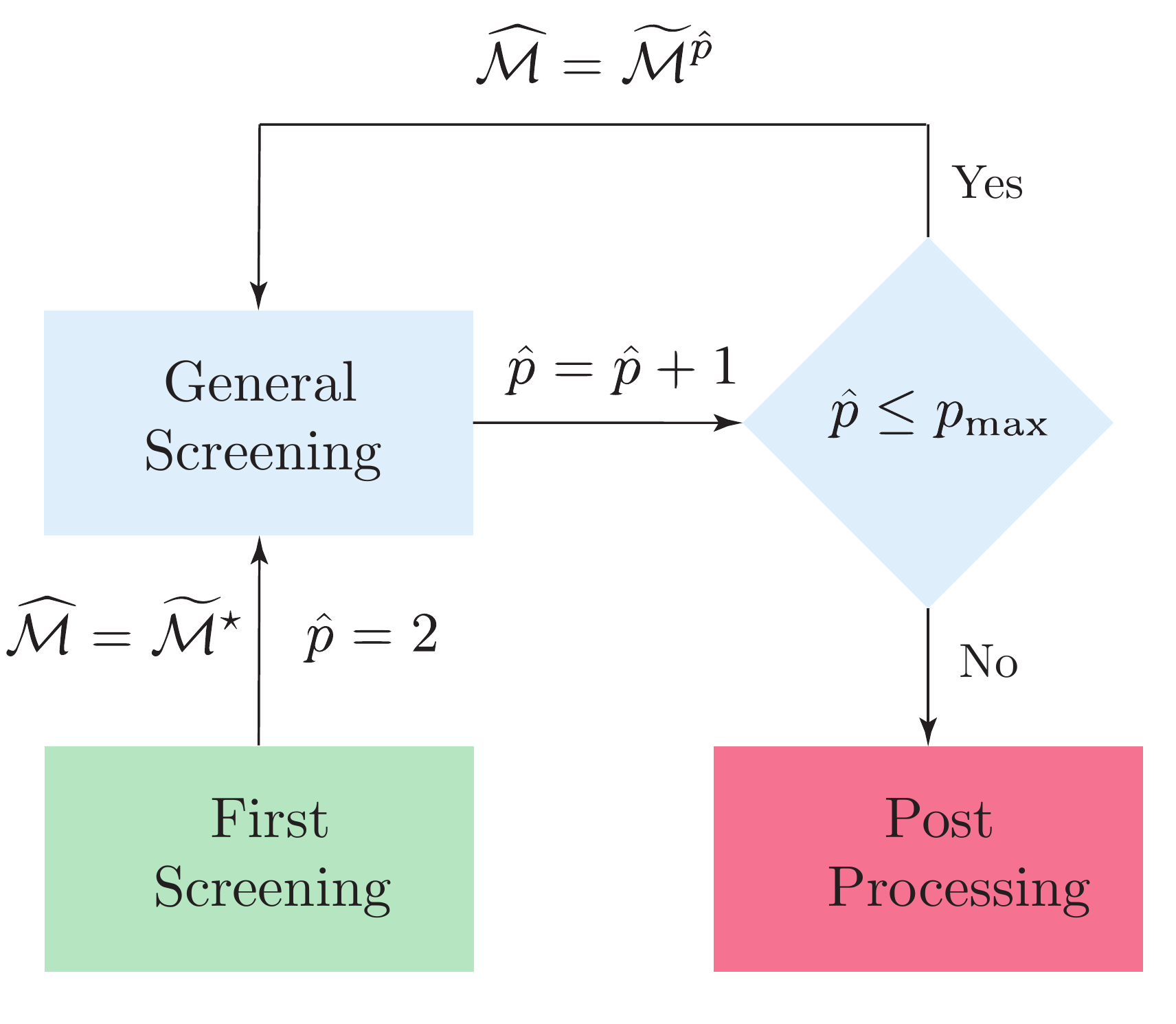}
\end{minipage}
\hspace{.3cm}
\begin{minipage}[c]{0.475\linewidth}
\centering
\begin{algorithm}[H]\small
\caption{SWAG}
\scriptsize
  \vspace{0.1cm}
INPUTS: A response $\mathbf{y} \in \mathbb{R}^n$ and attributes $\mathbf{X} \in \mathbb{R}^{n \times p}$; An attribute index set $\mathcal{S} := \{1, \hdots, p\}$; A learning mechanism $\mathcal{L}$; A maximum number of attributes $p_{\text{max}}$ ($< p$); A maximum number of learners $m$ for each step of the algorithm; A performance percentile $\alpha \in (0,1)$; the number of repetitions $r$ and the number of folds $k$ to compute the cross-validation training error. \\
\begin{algorithmic}[1]
  \setlength\itemsep{0em}
\STATE Run Algo. \ref{algo.firstscreen} using inputs $\mathbf{y}$, $\mathbf{X}$, $\mathcal{S}$, $\mathcal{L}$, $m$, $\alpha$ and obtain $\mathcal{S}^\star$ and $\widetilde{\mathcal{M}}^\star$
\STATE $\widehat{\mathcal{M}} \gets \widetilde{\mathcal{M}}^\star$
\STATE $\hat{p} \gets 2$
\WHILE {$\hat{p} \leq p_{\text{max}}$} 
        \STATE Run Algo. \ref{algo.generalscreen} using inputs $\mathbf{y}$, $\mathbf{X}$, $\mathcal{S}^\star$, $\widehat{\mathcal{M}}$, $\mathcal{L}$, $m$, $\alpha$ and obtain $\widetilde{\mathcal{M}}^{\hat{p}}$
        \STATE $\widehat{\mathcal{M}} \gets \widetilde{\mathcal{M}}^{\hat{p}}$
        \STATE $\hat{p} \gets \hat{p} + 1$
\ENDWHILE
\STATE Create
\setlength{\leftmargini}{0.2cm}
\begin{itemize}
    \item a set of strong learner sets $\widetilde{\mathcal{M}} := \{\widetilde{\mathcal{M}}^\star, \widetilde{\mathcal{M}}^2, \hdots, \widetilde{\mathcal{M}}^{p_{\text{max}}}\}$
    \item a set of training error vectors $\tilde{\epsilon} := \{\epsilon^\star, \epsilon^2, \hdots, \epsilon^{p_{\text{max}}}\}$
\end{itemize}

\end{algorithmic}
  \vspace{0.3cm}
OUTPUTS: $\widetilde{\mathcal{M}}$; $\tilde{\epsilon}$
\label{algo.seer}
\end{algorithm}
\end{minipage}\hfill
\end{figure*}

As \textit{per} its name, the above described algorithms provide a straightforward manner of selecting a set of predictive learners for a given attribute dimension $\hat{p}$. However, as the attribute dimension increases, the number of possible distinct attribute combinations increases exponentially fast. Therefore, as the attribute dimension grows, there is an increased risk of inefficiently exploring the attribute space if one simply randomly picks $m$ attribute combinations. For this reason, the SWAG performs a \textit{greedy} procedure which uses the information from Algo.~\ref{algo.firstscreen} to obtain the set of best attributes $\mathcal{S}^\star$ which is the easiest to explore completely. The next step of the algorithm then takes the set $\mathcal{S}^\star$ and the set of best learners $\widetilde{\mathcal{M}}^\star$ as the input $\widehat{\mathcal{M}}$ for Algo.~\ref{algo.generalscreen} for attribute dimension $\hat{p} = 2$. At each of the following steps $\hat{p} > 2$, the algorithm defines $\widehat{\mathcal{M}} := \widetilde{\mathcal{M}}^{\hat{p} - 1}$. Therefore, when increasing the attribute dimension, the algorithm only considers attribute combinations based on ``informative'' learners and attributes from the previous dimension. This procedure is repeated for all attribute dimensions until the maximal dimension $p_{\text{max}}$ is reached. Throughout the procedure, the algorithm saves the strong learner sets (whose training error is below the computed quantile) and training error vectors for each dimension $\hat{p}$. This procedure defines the SWAG which is described in Algo.~\ref{algo.seer}. Once the SWAG is run, there are a series of options that can be considered as discussed in the following paragraphs.

\paragraph{Post-Processing}
The output of the SWAG is a set of learners with high predictive power for each dimension of attribute combinations smaller or equal to $p_{\text{max}}$. The user could choose to directly make use of this set to perform predictions based on different sets of attributes or arrange attributes into networks for interpretation and exploration. However, these sets of learners could undergo an additional screening procedure according to the needs of the user. For example, an approach that is used for the applications in Sec.~\ref{sec.application} is to compute the median training error for each vector in the set $\tilde{\epsilon}$ (as defined in Algo.~\ref{algo.seer}) and select the quantile $\tilde{q}_{\delta}$ corresponding to the dimension whose median is the lowest, where $\delta \in (0,1)$ can differ from $\alpha$ (a possible choice is $0.01$). Having identified this quantile, the user can then select the learners (with the desired dimensions) whose training error is smaller or equal to this quantile.

\paragraph{Computational Complexity} 
The SWAG algorithm is a computationally intense procedure since it builds at most $p + m \; (p_{\text{max}}-1)$ learners in total which implies that the order of computation for any learner is multiplied by this factor. In addition, the $r$-repeated $k$-fold cross-validation will add complexity proportionally to the values of $r$ and $k$. Nevertheless, it must be recalled that the learners are built on an extremely small number of attributes (which implies computational efficiency for those mechanisms whose complexity highly depends on the number of attributes) and each fitting and cross-validation can be performed in parallel. A more formal study of the complexity of the SWAG and its possible improvement is left for further research.

\paragraph{Ensemble Learning}
While the set $\widetilde{\mathcal{M}}$ of strong low-dimensional learners can be used individually to obtain accurate predictions in different settings (where, for example, different attributes are available) or collectively to generate  interpretable networks, this set could eventually also be used within an ensemble approach. For example the SWAG learners could be included in Bagging \citep{breiman1996bagging}, Boosting \citep{schapire1990strength} or other model-averaging approaches (see e.g. \citep{raftery1997bayesian}) in order to possibly obtain more stable and accurate predictions. %In this sense, Sec.~\ref{sec.application} includes examples of a ``majority-rule'' averaging approach for classification problems using the set $\widetilde{\mathcal{M}}$.

\subsection{Meta-Learning Options}
\label{sec.meta_param}

For the application of the SWAG, we assume the user has chosen a learning mechanism $\mathcal{L}$ for which they would like would like to obtain more interpretable and replicable learners without losing much predictive power. Based on this, the user has to define the meta-parameters of the algorithm and eventually choose an ensemble approach to aggregate the strong learners that are output.

\paragraph{Meta-Parameters}
The main user-defined parameters for the SWAG are (i) the maximum attribute dimension $p_{\text{max}}$; (ii) the maximum number of learners $m$ to be built within each step and (iii) the performance percentile $\alpha$. Ideally, with unlimited computing power, the first two parameters would be as large as possible, i.e. $p_{\text{max}} = p$ and $m = \binom{p}{\lceil \frac{p}{2}\rceil}$. However, this defeats the purpose of the algorithm and therefore the decision of these parameters must be based mainly on interpretability/replicability requirements as well as available computing power and time. Below are some rules-of-thumb for the choice of these parameters:
\begin{itemize}
    \item $p_{\text{max}}$: Fixing the available computing power and the efficiency of the learning mechanism $\mathcal{L}$, this parameter will depend on the total dimension of the problem $p$. Indeed, the goal of the SWAG is to find extremely small dimensional learners. Therefore, even with very large $p$, one could always fix this parameter within a range of 5-20 (or smaller) for interpretability and/or replicability purposes. In addition, if a sparse (embedded) method is computationally efficient to run on the entire dataset, this parameter could be set to the number of selected attributes through this method (given computational constraints). Another criterion, when working with binary classification problems, is to use the \textit{Event Per Variable} (EPV) rule presented in \cite{vittinghoff2007relaxing} and discussed in \cite{van2014modern} (see Sec. \ref{sec.application} for example). In future work, this parameter can be implicitly determined by the algorithm based on the training error quantile (or other metric) thereby defining $p_{\text{max}}$ as the attribute dimension where the training error curve stops decreasing significantly similarly to the scree plot in factor or principal component analysis (see e.g. \citep{cattell1966scree}).
    \item $m$: Fixing the available computing power and the efficiency of the learning mechanism $\mathcal{L}$, this parameter will determine the proportion of attribute space that will be explored by the algorithm. We know that it depends on the size of the problem $p$ since we necessarily have $m \geq p$ for the screening step of Algo.~\ref{algo.firstscreen}. In addition, this parameter needs to be chosen considering the performance percentile $\alpha$: if $m$ is small and $\alpha$ is small, then the number of strong learners being selected could be extremely low (possibly zero). In general, we would want a large $m$ (so that $\alpha$ can eventually be chosen to be very small) and, remembering that $p^\star$ is the number of attributes released from Algo.~\ref{algo.firstscreen}, a rule-of-thumb is to set $m = \binom{p^\star}{2}$ (or close to it) in order to explore the entire (or most of the) subspace of two-dimensional learners generated by $p^\star$. 
    \item $\alpha$: as discussed in the previous point, this parameter is related to the maximum number of learners $m$. The larger $\alpha$, the more the attribute space is explored. Ideally, we want to choose a small $\alpha$ since we would want to select strong learners (with extremely low training error) and this is possible if $m$ is large enough. Generally good values for $\alpha$ are 0.01 or 0.05, implying that (roughly) 1\% or 5\% of the $m$ learners are selected at each step.
\end{itemize}

\section{Empirical Results}
\label{sec.application}

We study the empirical performance of the SWAG with different learning mechanisms and on different datasets taken from the UCI Machine Learning Repository \citep[see][]{Dua:2019}, ArrayExpress \citep[see][]{kolesnikov2015arrayexpress} and from the GitHub repository \url{https://github.com/ramhiser/datamicroarray}. More specifically, the chosen learning mechanisms are: (i) Lasso (logistic) \citep{friedman2010regularization}; (ii) Linear-Kernel Support Vector Machine (L-SVM) \citep[see e.g.][]{vapnik2013nature}; (iii)  Radial-Kernel SVM (R-SVM) \citep{cortes1995support}; (iv) Random Forest (RF)  \citep{breiman2001random}. To ensure a fair comparison, we run all the analyses with the \texttt{caret} R package \citep[see e.g.][for a review]{kuhn2008building}. All the hyper-parameters specific to each learning mechanism are set based on common/default choices and are discussed in the supplementary materials. With regards to the datasets, they are the following: (i) MeterA \citep{gyamfi2018linear}; (ii) LSVT  \citep{tsanas2013objective}; (iii) Ahus \citep{haakensen2016subtype}; (iv) Colon \citep{alon1999broad}; (v) Leukemia \citep{golub1999molecular}. The details regarding these data and the choices for the SWAG meta-learning options are listed in Tab.~\ref{tab_data}.

\begin{figure}[!ht]
\centering
\includegraphics[width=\linewidth]{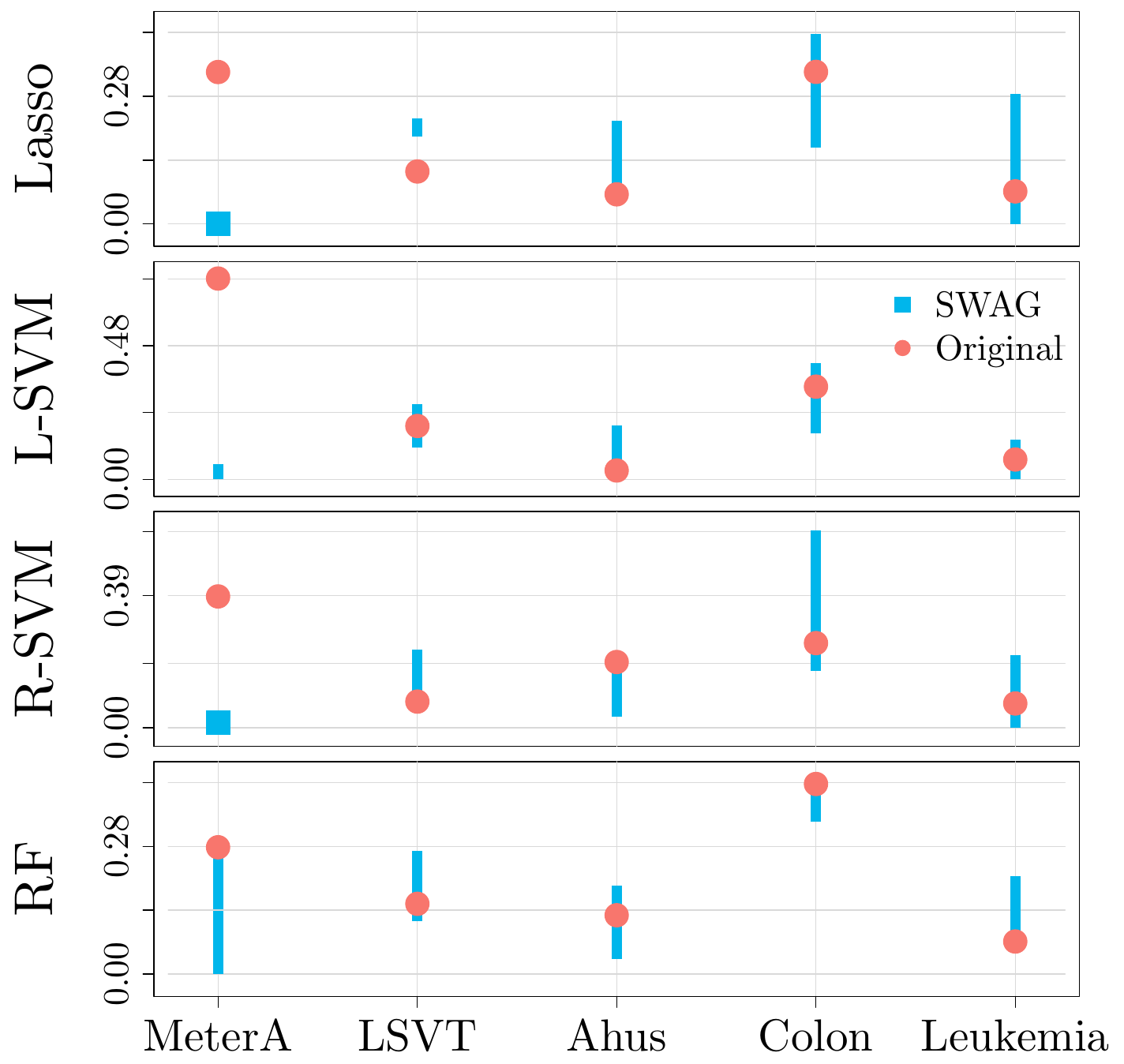}
\caption{SWAG test error range $\epsilon_{\text{test}}$ (blue rectangles) and original learning method test error (red dots). The blue \textit{squares} imply that the SWAG learners all have the same test error.}
\label{fig_epsilon_test}
\end{figure}

The choice of the SWAG meta-learning parameters is made in line with the rules of thumb described earlier and with the intent of reducing overall computational time while ensuring a reasonable exploration of the attribute space. More specifically, we choose $0.01 \leq \alpha \leq 0.05$ as a good compromise between fixing an $\alpha$ as small as possible (in order to select strong learners) and the possibility of exploring a considerable portion of the attribute space (see Sec. \ref{sec.meta_param}). With $m$ depending on the computing time and power available, we choose $m$ (as a linear function of $p_{\text{max}}$) in order to at least reasonably explore the subspace of two-dimensional learners for all the learning mechanisms and datasets (further discussed in supplementary material). The choice of $p_{\text{max}}$ is made guided by the EPV rule discussed in \cite{van2014modern}. A discussion on the sensitivity of the SWAG results with respect to the choice of these meta-parameters can be found in the supplementary material where it can be seen that the outcome is reasonably stable across different values of these parameters. Finally, we choose to apply 10-repeated 10-fold cross-validation (i.e. $r = 10$ and $k = 10$) within the SWAG and choose $\delta = 0.01$ for the post-processing based on the previously described median rule.
\begin{table}[!ht]
\fontsize{9.5pt}{9.5pt}
\centering
\caption{For each dataset: \# of instances per training and test set, \# of attributes and relative SWAG meta-parameter choices.}
\begin{tabular}{@{}lcccccc@{}}
\toprule
  Data & $n_{\text{train}}$ & $n_{\text{test}}$ & $p$  & $p_{\text{max}}$ & $m$ & $\alpha$  \\
 \cmidrule(r){1-1} \cmidrule(lr){2-2} \cmidrule(lr){3-3} \cmidrule(lr){4-4}  \cmidrule(lr){5-5}  \cmidrule(l){6-6}  \cmidrule(l){7-7}  
 MeterA  & 69 & 18 &  666  & 6 &  3984  & 0.05  \\
 LSVT & 100 & 26 &  310  & 6 &  1014  &  0.05   \\ %0.01 old
 Ahus & 125 & 31 &  15,739  & 8 &  99080   &   0.01   \\
Colon  & 50 & 12 & 2,000   & 4 &  7996  & 0.03 \\
Leukemia & 58 & 14 & 7,129 & 4 & 10160 &  0.01   \\
\bottomrule 
\end{tabular}
\label{tab_data}
\end{table}
With the above in mind, we present some summary measures of the set of SWAG learners $\widetilde{\mathcal{M}}$ such as their dimensions and their Jaccard diversity indices in order to understand how interpretable (sparse) and how replicable (diverse) the resulting SWAG networks are (see dataset-specific paragraphs further on). Tab. \ref{tab_jindex} collects this information for all the datasets where, for each dataset and learning mechanism, we provide the range of dimensions of the SWAG learners ($|s_l|$), the median value for the pairwise Jaccard indices for all learners in $\widetilde{\mathcal{M}}$ ($\text{med}_J$) as well as their range ($\text{range}_J$). It can be seen how the SWAG learners have all very low dimensions (also as a result of $p_\text{max}$) and that at least half of the Jaccard indices for all learning mechanisms and datasets are below or equal to 0.5, indicating that there is a reasonable or high level of diversity in the SWAG learners which is essential for replicability as well as interpretability. However, an important aspect to underline is that, while the goal of the SWAG is to deliver the latter two advantages, it comes at no (or extremely low) cost in terms of loss of prediction accuracy. Indeed, Fig.~\ref{fig_epsilon_test} represents the test error (denoted as $\epsilon_{test}$) of the original learning methods (i.e. learning methods applied to all attributes), represented by the red dots, and of the respective SWAG learners, represented by a blue rectangle proportional to the range of test errors (a similar figure is available for the training error $\epsilon_{train}$ in the supplementary material). It can be seen how in the majority of cases the SWAG learners have close or comparable (if not sometimes better) prediction accuracy with respect to their original versions, with homogeneous and exact accuracy when using Lasso and R-SVM on the Meter A data for example. Hence, the prediction performance is overall preserved while selecting learners of extremely low-dimension (maximum dimension 8) while the Lasso (the only sparse alternative considered) selects between 10 and 26 attributes. 

\begin{table*}[!ht]
\fontsize{8.5pt}{8.5pt}
\centering
\caption{For each learning mechanism (main columns) and each dataset (rows), we have the range of SWAG learner dimensions, the median Jaccard index and the range of the Jaccard indices for the SWAG learners.}
\begin{tabular}{@{}lcccccccccccc@{}}
\toprule
& \multicolumn{3}{c}{Lasso} & \multicolumn{3}{c}{L-SVM} & \multicolumn{3}{c}{R-SVM} & \multicolumn{3}{c}{RF} \\
 \cmidrule(lr){2-4} \cmidrule(lr){5-7}  \cmidrule(lr){8-10} \cmidrule(l){11-13}  
  Learner & $|s_l|$ & $\text{med}_J$ & $\text{range}_J$  & $|s_l|$ & $\text{med}_J$ & $\text{range}_J$ & $|s_l|$ & $\text{med}_J$ & $\text{range}_J$  & $|s_l|$ & $\text{med}_J$ & $\text{range}_J$  \\
 \cmidrule(r){1-1} \cmidrule(lr){2-2} \cmidrule(lr){3-3} \cmidrule(lr){4-4}  \cmidrule(lr){5-5}  \cmidrule(lr){6-6}  \cmidrule(lr){7-7} \cmidrule(lr){8-8} \cmidrule(lr){9-9} \cmidrule(lr){10-10} \cmidrule(lr){11-11}  \cmidrule(lr){12-12}  \cmidrule(l){13-13}  
 MeterA & $[4, 6]$ & $0.38$ & $[0.09,0.83]$  & $[2, 6]$ & $0.16$ & $[0.00,0.83]$ & $[4, 6]$ & $0.20$ & $[0.09,0.83]$  & $[4, 6]$ & $0.33$ & $[0.09,0.83]$   \\
LSVT  & $[4, 6]$ & $0.15$ & $[0.20,0.83]$ & $[5, 6]$ & $0.09$ & $[0.00,0.83]$  & $[4, 6]$ & $0.38$ & $[0.33,0.83]$ & $[3, 6]$ & $0.25$ & $[0.00,0.83]$  \\
 Ahus  & $[6, 8]$ & $0.15$ & $[0.00,0.88]$  & $[7, 8]$ & $0.23$ & $[0.00,0.88]$ & $[5, 8]$ & $0.23$ & $[0.00,0.75]$ & $[5, 8]$ & $0.25$ & $[0.00,0.75]$   \\
Colon  & $[3, 4]$ & $0.14$ & $[0.00,0.75]$  & $[3, 4]$ & $0.33$ & $[0.00,0.75]$ & $[3, 4]$ & $0.14$ & $[0.00,0.75]$ & $[3, 4]$ & $0.40$ & $[0.33,0.75]$ \\
Leukemia & $[2, 3]$ & $0.00$ & $[0.00,0.67]$ & $[2, 3]$ & $0.20$ & $[0.00,0.67]$  & $[2, 3]$ &$0.00$ & $[0.00,0.67]$  & $3$ &$0.50$ & $[0.20,0.50]$  \\
\bottomrule
\end{tabular}
\label{tab_jindex}
\end{table*} 

Given the fact that prediction accuracy is generally preserved with the SWAG learners, let us investigate the advantages of the SWAG in terms of its main goal which is interpretability (along with replicability which was observed through the Jaccard index above). To do so we consider building low-dimensional SWAG networks, based on the final selected set of learners, and give a brief overview of their advantages for research and interpretation.

\paragraph{Ultrasonic Flowmeters}

The Meter A data is analyzed in \cite{gyamfi2018linear} and collects measurements on ultrasonic flowmeter diagnostics. Achieving good diagnostics regarding the health of a flowmeter is of extreme importance for condition-based maintenance in many industrial sectors such as the oil and gas industry since incorrect measurement can entail considerable economic and material losses \citep[see e.g.][]{flowmeter}. In this data the goal is to classify the health of a meter into two classes: ``Healthy'' (Class 1) or ``Installation effects'' (Class 2). Given that the attributes are measurements of physical nature, we decide to consider all first-order interactions which finally delivers a total attribute size of $p = 666$ (36 original attributes plus $\binom{36}{2} = 630$ interactions).

\begin{figure}[!ht]
\centering
\includegraphics[scale =.68]{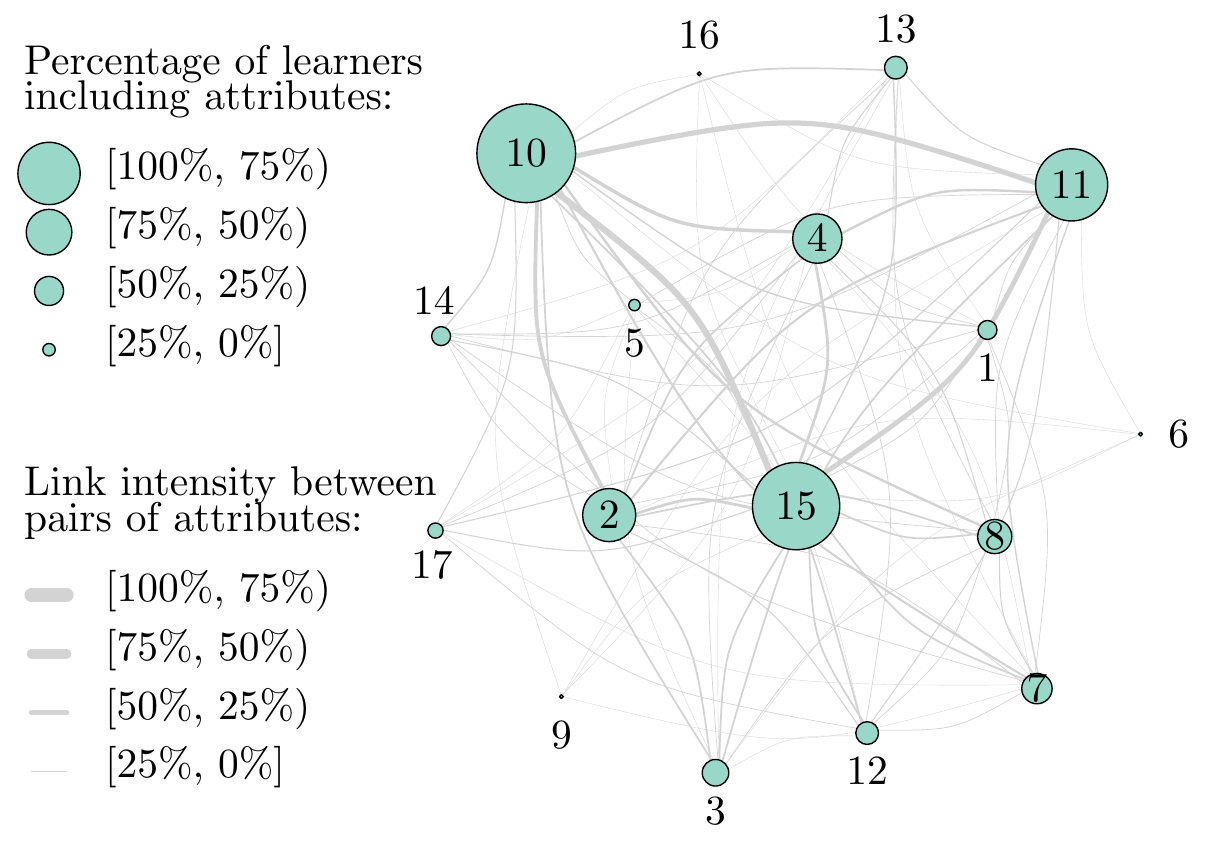}
\caption{Network of attribute importance and pairwise link-intensity in the set of learners $\widetilde{\mathcal{M}}$ for the Meter A dataset using the Lasso-based SWAG results.}
\label{fig:network}
\end{figure}

Using the results from Lasso-based SWAG, we can see how the attributes most frequently included in the selected learners (listed in supplementary material) can be arranged into a network where the edges represent the most common connections between these attributes as can be seen in Fig. \ref{fig:network}. Therefore, in order to understand the mechanics and perform diagnostics for this ultrasonic flowmeter, among the 666 attributes, a researcher could for example focus their attention on attributes labeled 10, 11 and 15 which correspond to the interaction (i) between flatness ratio and gain as well as (ii) between the speed of sound and gain at the first end (of the fifth path). As a final note, given that cost-based maintenance can have asymmetric costs according to the decision taken on the flowmeter, the SWAG could allow to select learners based on the corresponding (non-convex) cost-function instead of a symmetric kind of loss (i.e. each type of error is weighted equally) that learning mechanisms are usually based on \citep[see e.g.][]{crone2002training, masnadi2007asymmetric}.

\paragraph{LSVT}

The voice rehabilitation dataset was analyzed in \cite{tsanas2013objective} in order to assess the effectiveness of a computer program called ``Lee Silverman voice treatment (LSVT) Companion'' which allows patients with Parkinson's disease to independently progress through a rehabilitative treatment session. Taking data on 126 samples from 14 patients who followed the latter treatment, 310 dysphonia measures were taken on each of them (plus information on sex and age of the patients) and used to understand if they could correctly predict whether the patients' voices were ``acceptable'' or ``unacceptable'' after this treatment. In their analysis, a robust feature selection was used to select 8 attributes (based on the first eight attributes classified by the feature selection method) and subsequently R-SVM was tested (along with RF) to obtain around 90\% accuracy in classifying patients' progress.

\begin{figure}[!ht]
\centering
\includegraphics[scale =.68]{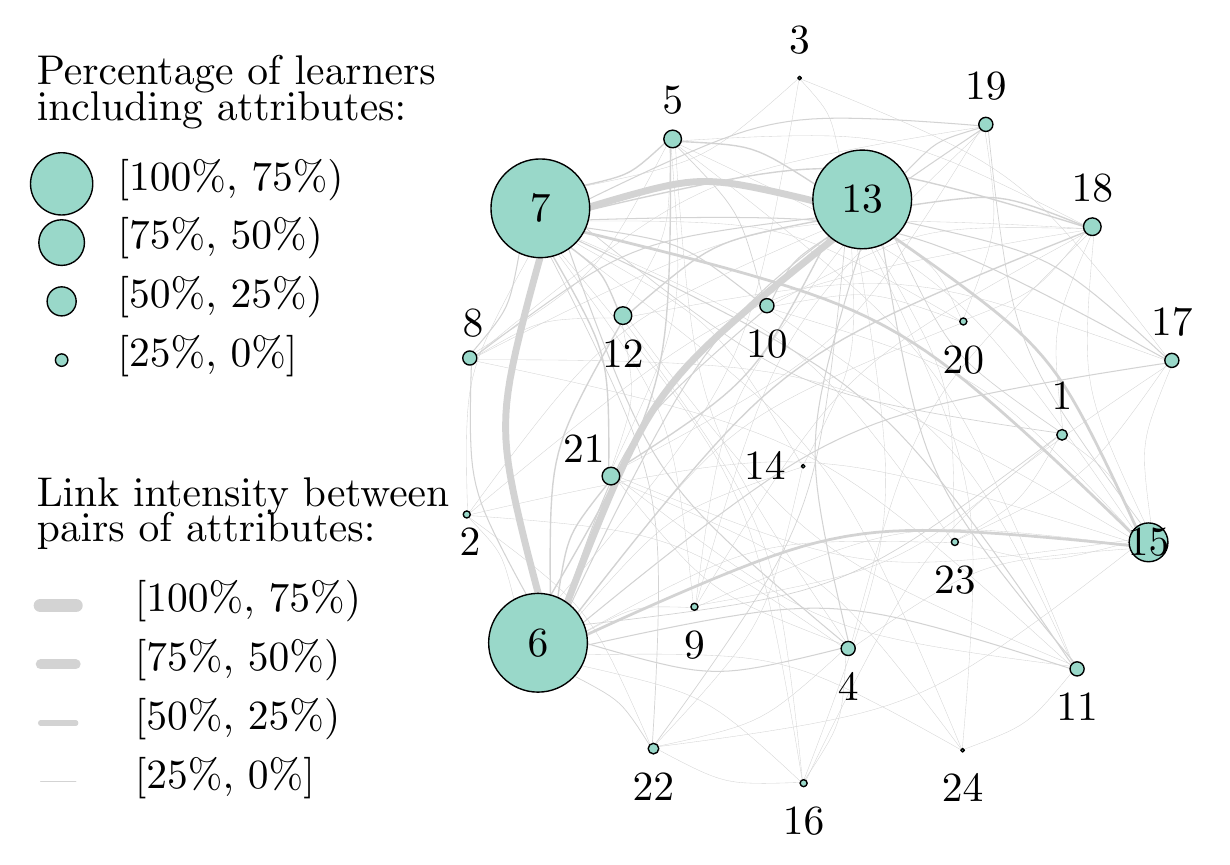}
\caption{Network of attribute importance and pairwise link-intensity in the set of learners $\widetilde{\mathcal{M}}$ for the LSVT dataset using the Radial-Kernel SVM SWAG results.}
\label{fig:network_lsvt}
\end{figure}

There is also scientific interest in determining the attributes (and combinations thereof) that most contribute to the definition, in this case, of a Parkinson's speech treatment as being acceptable or not. Also in this case, the SWAG learners (based on the R-SVM) can be arranged into a network in order to allow for interpretation as seen in Fig. \ref{fig:network_lsvt}. Based on the SWAG network, researchers interested in improving speech treatment should focus on attributes 6, 7 and 13 which correspond to the $2^{nd}$ and $3^{rd}$ Mel-Frequency Cepstral coefficients and on the entropy with base-4 logarithmic coefficients (as well as the interactions between these three attributes as highlighted by their frequent presence in the same SWAG learners). More information can be found in the supplementary material.

\paragraph{AHUS - Breast Cancer}

\begin{figure}[!ht]
\centering
\includegraphics[width =1\linewidth]{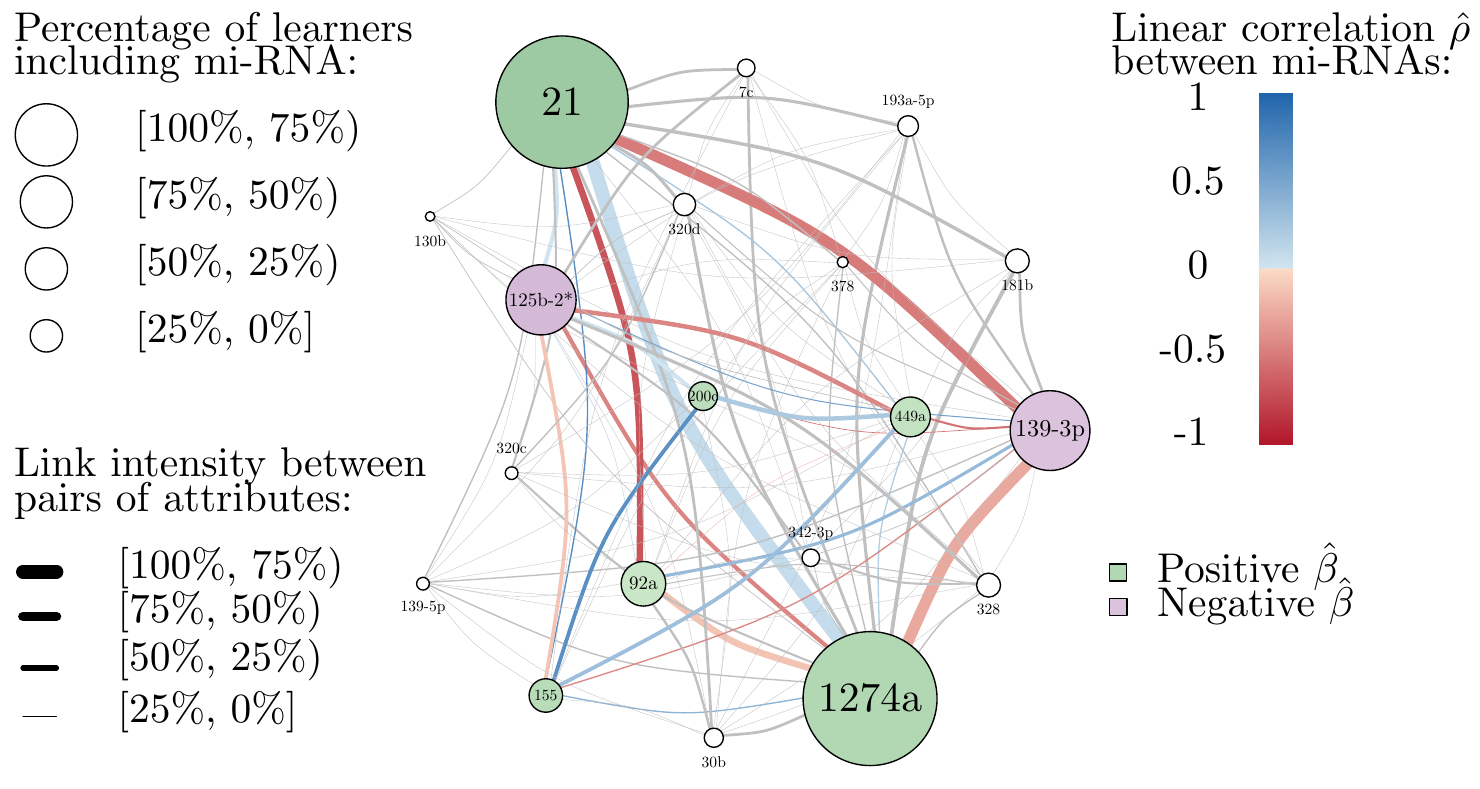}
\caption{Network of attribute importance and pairwise link-intensity in the set of learners $\widetilde{\mathcal{M}}$ for the AHUS dataset using the Lasso-based SWAG results.}
\label{fig:network_ahus}
\end{figure}

The Ahus breast cancer data is analyzed in \cite{haakensen2016subtype} and collects breast tissue specimens from patients over a period of 7 years and are profiled according to invasive carcinomas or normal breast tissue samples. Among others, the interest of this data is to understand how genetic biomarkers interact to explain the presence of breast cancer and, also, how these biomarkers can change their impact according to the attributes with which they interact \citep[see e.g.][]{miglioli2020breast}. In this sense, the SWAG network represented in Fig.~\ref{fig:network_ahus} can allow to visualize and determine the size and direction of impact of the attributes according to the other attributes they are connected to.

Indeed, having used Lasso, we can highlight additional features of the SWAG network such as the sign of the coefficients associated to the selected attributes as well as measures of correlation between connected attributes. For example, it can be seen how the \texttt{hsa-miR-1274a} and \texttt{hsa-miR-21} miRNAs appear to increase the probability of observing breast cancer and, although not being as frequently connected as with other attributes, have a positive correlation with each other. On the other hand, these highly predictive miRNAs are, for instance, frequently connected with \texttt{hsa-miR-139-3p} and \texttt{hsa-miR-92a} miRNAs with whom, on the contrary, they have a highly negative correlation. All these network features are extremely useful to respond to the above research questions and provide further insight for future directions of research in breast cancer genomics and the notion of ``chameleon'' miRNAs in systems biology \citep[see e.g.][]{stepanenko2013antagonistic}.

\section*{Discussion and Broader Impact}

A feature that could be interesting to investigate further for the SWAG is its possible capacity of selecting learners with fewer redundant attributes (e.g. correlated attributes) with higher probability. In fact, based on its screening procedure, the SWAG selects learners that improve predictive performance and therefore, at each increasing step, reduces the probability of selecting learners that do not improve their performance by adding redundant attributes. The latter attributes will not add further information than that already provided by the attributes included in the learner sets from the previous steps, implying that the new ``redunant'' learners will probably not be selected in the screening procedure.

In terms of practical consequences, an immediately noticeable impact is that the SWAG provides a reasonable approach to assess the prediction uncertainty of a given learner and also evaluate the performance of any learner compared to the distribution of its training and/or test errors. Indeed, if one does not require a low-dimensional strong learner, then they can always use the SWAG error distribution to understand whether their chosen learning mechanism appears to be better or at least comparable. In this sense, the SWAG error distribution can provide a ``validation metric'', in the direction outlined by \citep{nadeau2000inference}, to justify the use of a particular learning mechanism for a given dataset (in the same way a goodness-of-fit measure is used in statistical inference).

% MDA (see below) :“For each tree, the prediction accuracy on the out-of-bag portion of the data is recorded. Then the same is done after permuting each predictorvariable.  The difference between the two accuracies are then averaged over all trees, and normalized by the standard error.

However, the main impact of the SWAG in our view lies in its library of strong low-dimensional learners. Indeed the algorithm can integrate the advantages of mechanisms such as Random Forest where the importance of attributes can be retrieved based on the Mean Decrease Impurity (MDI) or the Mean Decrease Accuracy (MDA) of each predictor in the forest \citep[see e.g.][]{ishwaran2007variable, louppe2013understanding}. Therefore, even learning mechanisms that are characterized by a ``black-box''-type of procedure (as for the R-SVM example) can be made interpretable through the use of the SWAG. Moreover, based on the concept of ``synonyms'', the multiplicity of strong learners delivered by the SWAG can allow to adequately respond to predictive needs even in cases where many attributes are not collected or stored. For example, in the fields of natural sciences and medicine not all the responses or measurements may be available and a library of learners that can provide accurate prediction can be used to pick the learner(s) that best suit(s) the available information. 

%In the latter fields it is often the case that machines and material used to collect measurements during experiments or tests can only collect or process certain types of information and therefore cannot make use of results from other research to adequately predict responses (e.g. presence or absence of tumors). With the SWAG it is possible not only to interpret biological information (through the creation of gene networks for example) but also increase the probability of being able to make correct diagnoses when only having a limited amount of genetic or other information (attributes) available.

%On the other hand, the SWAG delivers a set of strong learners that can be arranged into low-dimensional and interpretable networks, as seen earlier, where the most common attributes, similarly to stability selection algorithms \citep{meinshausen2010stability}, can be considered as the main hubs in a network while the attributes connected to these main hubs can either be secondary hubs or be considered as ``synonyms'' of the other attributes.

Finally, the nature of the SWAG lends itself to the selection of strong learners based on non-convex loss (cost) functions that are different from those used to optimize the learners themselves. Depending on the ``cost'' of each learner-based decision, the cross-validation error within the SWAG could be evaluated based on asymmetric (or complex) cost functions and finally select learners that perform best in terms of this metric \citep[see e.g.][]{ting2000comparative}. This can have impacts in many sectors, from machine-maintenance to patient treatments, where decision-making can often be characterized by asymmetric costs and the availability of limited or very specific sets of information.

\newpage
\bibliography{biblio}

\begin{thebibliography}{44}
\providecommand{\natexlab}[1]{#1}
\providecommand{\url}[1]{\texttt{#1}}
\expandafter\ifx\csname urlstyle\endcsname\relax
  \providecommand{\doi}[1]{doi: #1}\else
  \providecommand{\doi}{doi: \begingroup \urlstyle{rm}\Url}\fi

\bibitem[Alon et~al.(1999)Alon, Barkai, Notterman, Gish, Ybarra, Mack, and
  Levine]{alon1999broad}
Alon, U., Barkai, N., Notterman, D.~A., Gish, K., Ybarra, S., Mack, D., and
  Levine, A.~J.
\newblock Broad patterns of gene expression revealed by clustering analysis of
  tumor and normal colon tissues probed by oligonucleotide arrays.
\newblock \emph{Proceedings of the National Academy of Sciences}, 96\penalty0
  (12):\penalty0 6745--6750, 1999.

\bibitem[Anderson \& Burnham(2004)Anderson and Burnham]{anderson2004model}
Anderson, D. and Burnham, K.
\newblock Model selection and multi-model inference.
\newblock \emph{Second. NY: Springer-Verlag}, 63\penalty0 (2020):\penalty0 10,
  2004.

\bibitem[Breiman(1996)]{breiman1996bagging}
Breiman, L.
\newblock Bagging predictors.
\newblock \emph{Machine Learning}, 24\penalty0 (2):\penalty0 123--140, 1996.

\bibitem[Breiman(2001)]{breiman2001random}
Breiman, L.
\newblock Random forests.
\newblock \emph{Machine Learning}, 45\penalty0 (1):\penalty0 5--32, 2001.

\bibitem[Burnham \& Anderson(2004)Burnham and Anderson]{burnham2004multimodel}
Burnham, K.~P. and Anderson, D.~R.
\newblock Multimodel inference: understanding aic and bic in model selection.
\newblock \emph{Sociological methods \& research}, 33\penalty0 (2):\penalty0
  261--304, 2004.

\bibitem[Carmona-Cejudo et~al.(2011)Carmona-Cejudo, Baena-Garc{\'\i}a,
  Campo-Avila, Morales-Bueno, Gama, and Bifet]{carmona2011dataminingonline}
Carmona-Cejudo, J.~M., Baena-Garc{\'\i}a, M., Campo-Avila, J., Morales-Bueno,
  R., Gama, J., and Bifet, A.
\newblock Using gnusmail to compare data stream mining methods for on-line
  email classification.
\newblock In \emph{Proceedings of the Second Workshop on Applications of
  Pattern Analysis}, pp.\  12--18, 2011.

\bibitem[Caruana et~al.(2004)Caruana, Niculescu-Mizil, Crew, and
  Ksikes]{caruana2004ensemble}
Caruana, R., Niculescu-Mizil, A., Crew, G., and Ksikes, A.
\newblock Ensemble selection from libraries of models.
\newblock In \emph{Proceedings of the 21st International Conference on Machine
  Learning}, pp.\ ~18, 2004.

\bibitem[Cattell(1966)]{cattell1966scree}
Cattell, R.~B.
\newblock The scree test for the number of factors.
\newblock \emph{Multivariate Behavioral Research}, 1\penalty0 (2):\penalty0
  245--276, 1966.

\bibitem[Chandrashekar \& Sahin(2014)Chandrashekar and
  Sahin]{chandrashekar2014survey}
Chandrashekar, G. and Sahin, F.
\newblock A survey on feature selection methods.
\newblock \emph{Computers \& Electrical Engineering}, 40\penalty0 (1):\penalty0
  16--28, 2014.

\bibitem[Cortes \& Vapnik(1995)Cortes and Vapnik]{cortes1995support}
Cortes, C. and Vapnik, V.
\newblock Support vector machine.
\newblock \emph{Machine Learning}, 20\penalty0 (3):\penalty0 273--297, 1995.

\bibitem[Crone(2002)]{crone2002training}
Crone, S.~F.
\newblock Training artificial neural networks for time series prediction using
  asymmetric cost functions.
\newblock In \emph{Proceedings of the 9th International Conference on Neural
  Information Processing, 2002. ICONIP'02.}, volume~5, pp.\  2374--2380. IEEE,
  2002.

\bibitem[Draghici et~al.(2006)Draghici, Khatri, Eklund, and
  Szallasi]{draghici2006reliability}
Draghici, S., Khatri, P., Eklund, A.~C., and Szallasi, Z.
\newblock Reliability and reproducibility issues in dna microarray
  measurements.
\newblock \emph{TRENDS in Genetics}, 22\penalty0 (2):\penalty0 101--109, 2006.

\bibitem[Dua \& Graff(2017)Dua and Graff]{Dua:2019}
Dua, D. and Graff, C.
\newblock {UCI} machine learning repository, 2017.
\newblock URL \url{http://archive.ics.uci.edu/ml}.

\bibitem[Friedman et~al.(2010)Friedman, Hastie, and
  Tibshirani]{friedman2010regularization}
Friedman, J., Hastie, T., and Tibshirani, R.
\newblock Regularization paths for generalized linear models via coordinate
  descent.
\newblock \emph{Journal of Statistical Software}, 33\penalty0 (1):\penalty0 1,
  2010.

\bibitem[Golub et~al.(1999)Golub, Slonim, Tamayo, Huard, Gaasenbeek, Mesirov,
  Coller, Loh, Downing, Caligiuri, et~al.]{golub1999molecular}
Golub, T.~R., Slonim, D.~K., Tamayo, P., Huard, C., Gaasenbeek, M., Mesirov,
  J.~P., Coller, H., Loh, M.~L., Downing, J.~R., Caligiuri, M.~A., et~al.
\newblock Molecular classification of cancer: class discovery and class
  prediction by gene expression monitoring.
\newblock \emph{science}, 286\penalty0 (5439):\penalty0 531--537, 1999.

\bibitem[Guerrier et~al.(2016)Guerrier, Mili, Molinari, Orso, Avella-Medina,
  and Ma]{guerrier2016predictive}
Guerrier, S., Mili, N., Molinari, R., Orso, S., Avella-Medina, M., and Ma, Y.
\newblock A predictive based regression algorithm for gene network selection.
\newblock \emph{Frontiers in Genetics}, 7:\penalty0 97, 2016.

\bibitem[Gyamfi et~al.(2018)Gyamfi, Brusey, Hunt, and Gaura]{gyamfi2018linear}
Gyamfi, K.~S., Brusey, J., Hunt, A., and Gaura, E.
\newblock Linear dimensionality reduction for classification via a sequential
  bayes error minimisation with an application to flow meter diagnostics.
\newblock \emph{Expert Systems with Applications}, 91:\penalty0 252--262, 2018.

\bibitem[Haakensen et~al.(2016)Haakensen, Nygaard, Greger, Aure, Fromm,
  Bukholm, L{\"u}ders, Chin, Git, Caldas, et~al.]{haakensen2016subtype}
Haakensen, V.~D., Nygaard, V., Greger, L., Aure, M.~R., Fromm, B., Bukholm,
  I.~R., L{\"u}ders, T., Chin, S.-F., Git, A., Caldas, C., et~al.
\newblock Subtype-specific micro-rna expression signatures in breast cancer
  progression.
\newblock \emph{International Journal of Cancer}, 139\penalty0 (5):\penalty0
  1117--1128, 2016.

\bibitem[Harrison et~al.(2018)Harrison, Donaldson, Correa-Cano, Evans, Fisher,
  Goodwin, Robinson, Hodgson, and Inger]{harrison2018brief}
Harrison, X.~A., Donaldson, L., Correa-Cano, M.~E., Evans, J., Fisher, D.~N.,
  Goodwin, C.~E., Robinson, B.~S., Hodgson, D.~J., and Inger, R.
\newblock A brief introduction to mixed effects modelling and multi-model
  inference in ecology.
\newblock \emph{PeerJ}, 6:\penalty0 e4794, 2018.

\bibitem[Ishwaran(2007)]{ishwaran2007variable}
Ishwaran, H.
\newblock Variable importance in binary regression trees and forests.
\newblock \emph{Electronic Journal of Statistics}, 1:\penalty0 519--537, 2007.

\bibitem[Kolesnikov et~al.(2015)Kolesnikov, Hastings, Keays, Melnichuk, Tang,
  Williams, Dylag, Kurbatova, Brandizi, and
  Burdett]{kolesnikov2015arrayexpress}
Kolesnikov, N., Hastings, E., Keays, M., Melnichuk, O., Tang, Y.~A., Williams,
  E., Dylag, M., Kurbatova, N., Brandizi, M., and Burdett, T.
\newblock Arrayexpress update—simplifying data submissions.
\newblock \emph{Nucleic Acids Research}, 43\penalty0 (D1):\penalty0
  D1113--D1116, 2015.

\bibitem[Kuhn(2008)]{kuhn2008building}
Kuhn, M.
\newblock Building predictive models in r using the caret package.
\newblock \emph{Journal of Statistical Software}, 28\penalty0 (5):\penalty0
  1--26, 2008.

\bibitem[Louppe et~al.(2013)Louppe, Wehenkel, Sutera, and
  Geurts]{louppe2013understanding}
Louppe, G., Wehenkel, L., Sutera, A., and Geurts, P.
\newblock Understanding variable importances in forests of randomized trees.
\newblock In \emph{Advances in Neural Information Processing Systems}, pp.\
  431--439, 2013.

\bibitem[Marigorta et~al.(2018)Marigorta, Rodr{\'\i}guez, Gibson, and
  Navarro]{marigorta2018replicability}
Marigorta, U.~M., Rodr{\'\i}guez, J.~A., Gibson, G., and Navarro, A.
\newblock Replicability and prediction: lessons and challenges from gwas.
\newblock \emph{Trends in Genetics}, 34\penalty0 (7):\penalty0 504--517, 2018.

\bibitem[Masnadi-Shirazi \& Vasconcelos(2007)Masnadi-Shirazi and
  Vasconcelos]{masnadi2007asymmetric}
Masnadi-Shirazi, H. and Vasconcelos, N.
\newblock Asymmetric boosting.
\newblock In \emph{Proceedings of the 24th International Conference on Machine
  Learning}, pp.\  609--619, 2007.

\bibitem[Meinshausen \& Yu(2009)Meinshausen and Yu]{meinshausen2009lasso}
Meinshausen, N. and Yu, B.
\newblock Lasso-type recovery of sparse representations for high-dimensional
  data.
\newblock \emph{The Annals of Statistics}, 37\penalty0 (1):\penalty0 246--270,
  2009.

\bibitem[Miglioli et~al.(2020)Miglioli, Bakalli, Guerrier, Orso, Molinari, and
  Mili]{miglioli2020breast}
Miglioli, C., Bakalli, G., Guerrier, S., Orso, S., Molinari, R., and Mili, N.
\newblock Chameleon micrornas in breast cancer: their elusive role as
  regulatory factors in cancer progression.
\newblock \emph{https://doi.org/10.1101/2020.12.15.422846}, 2020.

\bibitem[Nadeau \& Bengio(2000)Nadeau and Bengio]{nadeau2000inference}
Nadeau, C. and Bengio, Y.
\newblock Inference for the generalization error.
\newblock In \emph{Advances in Neural Information Processing Systems}, pp.\
  307--313, 2000.

\bibitem[Quinn et~al.(2020)Quinn, Nguyen, Rana, Gupta, and
  Venkatesh]{quinn2020deepcoda}
Quinn, T., Nguyen, D., Rana, S., Gupta, S., and Venkatesh, S.
\newblock Deepcoda: personalized interpretability for compositional health
  data.
\newblock In \emph{International Conference on Machine Learning}, pp.\
  7877--7886. PMLR, 2020.

\bibitem[Raftery et~al.(1997)Raftery, Madigan, and
  Hoeting]{raftery1997bayesian}
Raftery, A.~E., Madigan, D., and Hoeting, J.~A.
\newblock Bayesian model averaging for linear regression models.
\newblock \emph{Journal of the American Statistical Association}, 92\penalty0
  (437):\penalty0 179--191, 1997.

\bibitem[Schapire(1990)]{schapire1990strength}
Schapire, R.~E.
\newblock The strength of weak learnability.
\newblock \emph{Machine Learning}, 5\penalty0 (2):\penalty0 197--227, 1990.

\bibitem[Stepanenko et~al.(2013)Stepanenko, Vassetzky, and
  Kavsan]{stepanenko2013antagonistic}
Stepanenko, A., Vassetzky, Y., and Kavsan, V.
\newblock Antagonistic functional duality of cancer genes.
\newblock \emph{Gene}, 529\penalty0 (2):\penalty0 199--207, 2013.

\bibitem[Ting(2000)]{ting2000comparative}
Ting, K.~M.
\newblock A comparative study of cost-sensitive boosting algorithms.
\newblock In \emph{Proceedings of the 17th International Conference on Machine
  Learning}. Citeseer, 2000.

\bibitem[Tsanas et~al.(2013)Tsanas, Little, Fox, and
  Ramig]{tsanas2013objective}
Tsanas, A., Little, M.~A., Fox, C., and Ramig, L.~O.
\newblock Objective automatic assessment of rehabilitative speech treatment in
  parkinson's disease.
\newblock \emph{IEEE Transactions on Neural Systems and Rehabilitation
  Engineering}, 22\penalty0 (1):\penalty0 181--190, 2013.

\bibitem[TUV-NEL(2012)]{flowmeter}
TUV-NEL.
\newblock Testing the diagnostic capabilities of liquid ultrasonic flow meters.
\newblock \emph{National Measurement System}, 2012.

\bibitem[van~der Ploeg et~al.(2014)van~der Ploeg, Austin, and
  Steyerberg]{van2014modern}
van~der Ploeg, T., Austin, P.~C., and Steyerberg, E.~W.
\newblock Modern modelling techniques are data hungry: a simulation study for
  predicting dichotomous endpoints.
\newblock \emph{BMC medical research methodology}, 14\penalty0 (1):\penalty0
  1--13, 2014.

\bibitem[Vapnik(2013)]{vapnik2013nature}
Vapnik, V.
\newblock \emph{The Nature of Statistical Learning Theory}.
\newblock Springer science \& business media, 2013.

\bibitem[Vats \& Baraniuk(2013)Vats and Baraniuk]{vats2013doubt}
Vats, D. and Baraniuk, R.
\newblock When in doubt, swap: High-dimensional sparse recovery from correlated
  measurements.
\newblock In \emph{Advances in Neural Information Processing Systems}, pp.\
  989--997, 2013.

\bibitem[Vaughan \& Chen(2015)Vaughan and Chen]{vaughan2015data}
Vaughan, L. and Chen, Y.
\newblock Data mining from web search queries: A comparison of google trends
  and baidu index.
\newblock \emph{Journal of the Association for Information Science and
  Technology}, 66\penalty0 (1):\penalty0 13--22, 2015.

\bibitem[Vittinghoff \& McCulloch(2007)Vittinghoff and
  McCulloch]{vittinghoff2007relaxing}
Vittinghoff, E. and McCulloch, C.~E.
\newblock Relaxing the rule of ten events per variable in logistic and cox
  regression.
\newblock \emph{American Journal of Epidemiology}, 165\penalty0 (6):\penalty0
  710--718, 2007.

\bibitem[Wang et~al.(2018)Wang, Li, Zuluaga, Pratt, Patel, Aertsen, Doel,
  David, Deprest, Ourselin, et~al.]{wang2018interactive}
Wang, G., Li, W., Zuluaga, M.~A., Pratt, R., Patel, P.~A., Aertsen, M., Doel,
  T., David, A.~L., Deprest, J., Ourselin, S., et~al.
\newblock Interactive medical image segmentation using deep learning with
  image-specific fine tuning.
\newblock \emph{IEEE Transactions on Medical Imaging}, 37\penalty0
  (7):\penalty0 1562--1573, 2018.

\bibitem[Wang(2019)]{wang2019gaining}
Wang, T.
\newblock Gaining free or low-cost interpretability with interpretable partial
  substitute.
\newblock In \emph{International Conference on Machine Learning}, pp.\
  6505--6514. PMLR, 2019.

\bibitem[Xiong et~al.(2001)Xiong, Fang, and Zhao]{xiong2001biomarker}
Xiong, M., Fang, X., and Zhao, J.
\newblock Biomarker identification by feature wrappers.
\newblock \emph{Genome Research}, 11\penalty0 (11):\penalty0 1878--1887, 2001.

\bibitem[Zhang et~al.(2015)Zhang, Xu, Yang, Li, and Zhang]{zhang2015survey}
Zhang, Z., Xu, Y., Yang, J., Li, X., and Zhang, D.
\newblock A survey of sparse representation: algorithms and applications.
\newblock \emph{IEEE Access}, 3:\penalty0 490--530, 2015.

\end{thebibliography}
\bibliographystyle{icml2021}

%\appendix
%\section{Do \emph{not} have an appendix here}

%\textbf{\emph{Do not put content after the references.}}
%
%Put anything that you might normally include after the references in a separate
%supplementary file.

%We recommend that you build supplementary material in a separate document.
%If you must create one PDF and cut it up, please be careful to use a tool that
%doesn't alter the margins, and that doesn't aggressively rewrite the PDF file.
%pdftk usually works fine. 

%\textbf{Please do not use Apple's preview to cut off supplementary material.} %In
%previous years it has altered margins, and created headaches at the camera-ready
%stage. 
%%%%%%%%%%%%%%%%%%%%%%%%%%%%%%%%%%%%%%%%%%%%%%%%%%%%%%%%%%%%%%%%%%%%%%%%%%%%%%%
%%%%%%%%%%%%%%%%%%%%%%%%%%%%%%%%%%%%%%%%%%%%%%%%%%%%%%%%%%%%%%%%%%%%%%%%%%%%%%%

\end{document}